\documentclass[10pt,twocolumn,letterpaper]{article}

\usepackage[pagenumbers]{iccv} 

\usepackage{amsmath}
\usepackage{amssymb}
\usepackage{xcolor}
\usepackage{booktabs}
\usepackage{multirow}
\usepackage{colortbl}

\usepackage{placeins}
\usepackage[linesnumbered,ruled,vlined]{algorithm2e}
\newtheorem{definition}{Definition}

\definecolor{darkpastelgreen}{rgb}{0.01, 0.75, 0.24}
\definecolor{darkgreen}{rgb}{0.00, 0.8, 0.2}
\definecolor{darkyellow}{rgb}{0.96, 0.75, 0.00}

\definecolor{iccvblue}{rgb}{0.21,0.49,0.74}
\usepackage[pagebackref,breaklinks,colorlinks,allcolors=iccvblue]{hyperref}

\def\paperID{4713} 
\def\confName{ICCV}
\def\confYear{2025}

\title{Interaction-Merged Motion Planning: Effectively Leveraging Diverse Motion Datasets for Robust Planning}

\author{
Giwon Lee$^{1*}$, Wooseong Jeong$^{1*}$, Daehee Park$^{2}$, Jaewoo Jeong$^{1}$, and Kuk-Jin Yoon$^{1}$ \\
$^1$Visual Intelligence Lab., KAIST, Korea \\
$^2$Intelligent Systems and Learning Lab., DGIST, Korea
}

\begin{document}
\maketitle

\renewcommand{\thefootnote}{\fnsymbol{footnote}}
\footnotetext[1]{Equal contribution to this work.\\
Our source code is available at: \url{https://github.com/wooseong97/IMMP}}

\begin{abstract}
Motion planning is a crucial component of autonomous robot driving. While various trajectory datasets exist, effectively utilizing them for a target domain remains challenging due to differences in agent interactions and environmental characteristics. Conventional approaches, such as domain adaptation or ensemble learning, leverage multiple source datasets but suffer from domain imbalance, catastrophic forgetting, and high computational costs. To address these challenges, we propose \textbf{Interaction-Merged Motion Planning (IMMP)}, a novel approach that leverages parameter checkpoints trained on different domains during adaptation to the target domain. IMMP follows a two-step process: pre-merging to capture agent behaviors and interactions, sufficiently extracting diverse information from the source domain, followed by merging to construct an adaptable model that efficiently transfers diverse interactions to the target domain. Our method is evaluated on various planning benchmarks and models, demonstrating superior performance compared to conventional approaches.
\end{abstract}    

\section{Introduction}
Motion planning~\cite{sadat2020perceive_intro_planning1,zeng2020dsdnet_intro_planning2,chen2024end_intro_planning3,yurtsever2020survey_intro_planning4} is the final step in autonomous robot driving, focused on generating safe and efficient trajectory based on perception and prediction. 
With the availability of various trajectory datasets~\cite{chen2019CrowdNav,pellegrini2009ETH,lerner2007UCY, rudenko2020THOR, bae2023SIT, yan2017onlineLCAS, martin2021JRDB, jeong2024multi}, data-driven approaches have become more popular due to their ability to adapt to diverse driving scenarios~\cite{wang2023dreamwalker_planning2, yang2024diffusion_planning1,huang2023gameformer,kedia2023GameTheoretic,huang2023DIPP,huang2024DTPP,jeong2025multi}.

However, existing trajectory datasets exhibit distinct characteristics, including differences in scene composition, agent interaction tendencies, and the difficulties in data acquisition. 
Indoor datasets are constrained in movement due to static structures, where robots generally move at slower speeds. 
In contrast, outdoor datasets encompass unconstrained environments and results in a more dynamic maneuver.
To account for these diverse environments, Human-Human Interaction (HHI) datasets and Human-Robot Interaction (HRI) datasets are collectively utilized to train a planner model.
However, using HHI datasets for planning assumes human locomotion as ego-agent motion, therefore incorporating interaction dynamics that differ from HRI datasets where real robot-human interactions shape agent behaviors.
These inherent differences across datasets collectively lead to variations in the nature of agents' maneuvers and their interactions.
Holistically accounting for such disparities is crucial towards enhancing the overall adaptability in diverse real-world scenarios.

\begin{figure}[t]
  \centering
   \includegraphics[width=\linewidth]{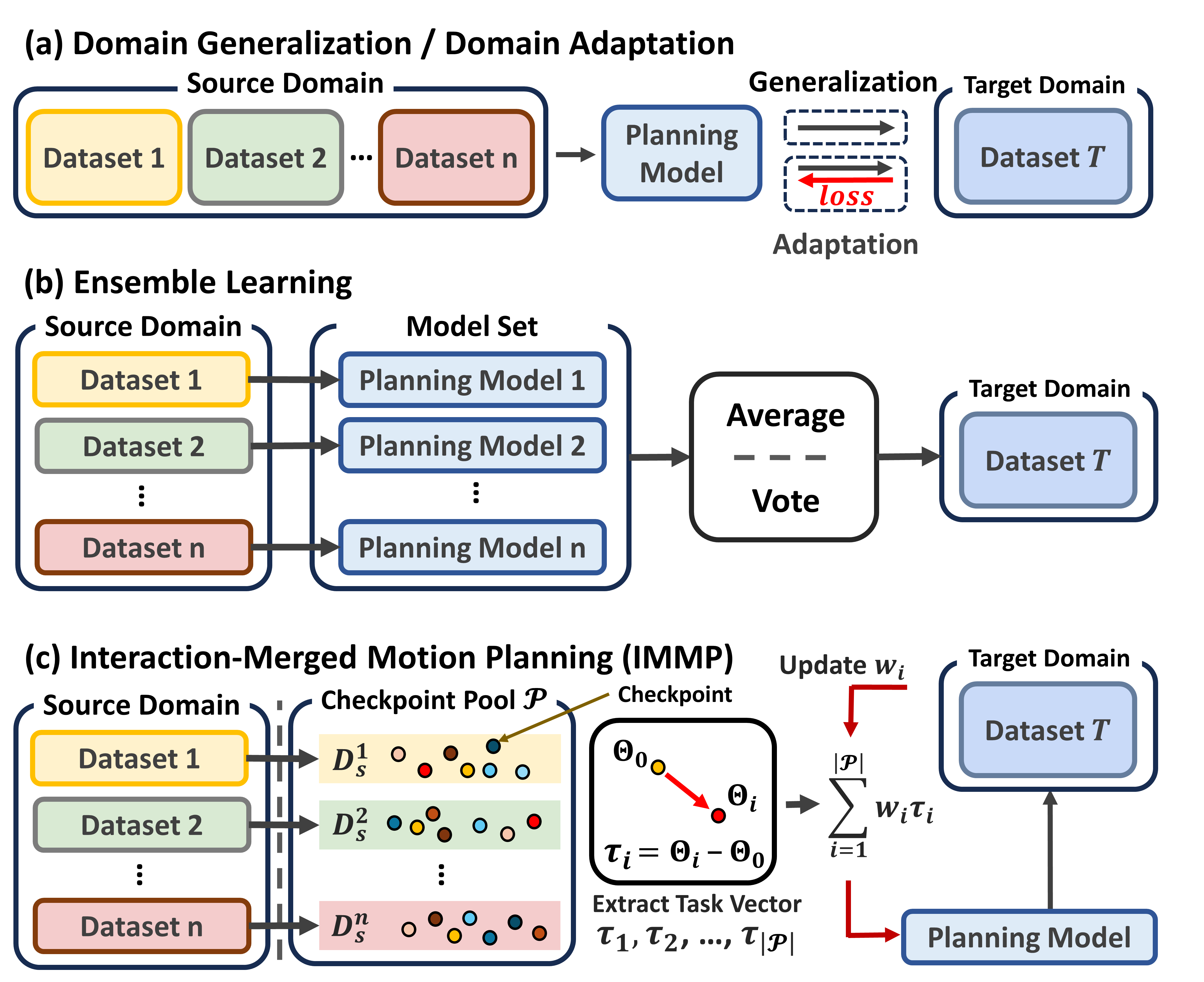}
   \caption{Comparison of different approaches for leveraging the source domains for target domain adaptation. (a) Domain generalization or adaptation directly utilizes relationships within datasets. (b) Ensemble learning combines predictions from multiple planning models, each trained on different datasets. (c) The proposed IMMP collects parameter checkpoints $\Theta$ and merges task vectors $\tau$, which capture differences between trained and initial parameters, to improve adaptation to the target domain.}
   \label{fig:viz_intro}
   \vspace{-5pt}
\end{figure}

One straightforward approach is to incorporate multiple source datasets into the training set (\cref{fig:viz_intro} (a)).
Training on multiple datasets acquired from disparate environments can leverage richer patterns compared to training on a single dataset~\cite{feng2024unitrajDomainGeneralization,zhu2024unitraj2DomainGeneralization}. 
However, naively increasing the number of source datasets fails to incorporate the inherent differences between datasets due to two aspects.
First, the most effective source datasets for improving performance on a given target domain are unknown before actually training and fine-tuning a model. 
Differences in dataset properties—such as agent behaviors, interaction patterns, and environment characteristics-can lead to conflicts, making direct dataset composition non-trivial. 
These disparities in dataset can cause what we refer to as the \textit{domain imbalance problem}, where certain datasets disproportionately affect the learning process, dominating model updates and affecting different modules or parameters within the planning model. 
Second, domain imbalance can lead to knowledge interference, similar to catastrophic forgetting in continual learning~\cite{kirkpatrick2017overcomingcatastrophicforgetting1,schwarz2018progresscatastrophicforgetting2,shin2017continualcatastrophicforgetting3,shmelkov2017incrementalcatastrophicforgetting4}, where newly introduced domains may override previously learned information. 

Another approach is to integrate multiple planner models (\cref{fig:viz_intro} (b)). 
Despite the improvement in generalization on different domains, ensemble approaches require multiple planner models during inference and therefore increases the computational requirement. 
These challenges suggest that simply aggregating diverse motion datasets or using multiple models are both complex and inefficient approaches towards enhancing motion planning adaptability.

In response, we leverage model merging~\cite{ortiz2024task, demir2024adaptive, wortsman2022model, wortsman2022robust, yadav2023ties} to integrate knowledge from multiple sources to improve generalization while retaining computational cost.
Specifically, model parameters from different checkpoints, each fine-tuned on a specific dataset, are merged. 
Each model's parameters encode dataset-specific motion patterns and interaction dynamics, and model merging effectively combines this multi-dataset knowledge to adapt to a new target dataset.
However, existing merging methods struggle in motion planning due to two key limitations. First, they fail to construct a sufficient parameter checkpoint pool, as they do not account for diverse evaluation metrics and multi-objective loss functions. 
Second, they overlook the hierarchical nature of motion planning features, potentially disrupting feature structures or discarding critical information.
To address these limitations, we propose a \textbf{motion-planning-targeted merging method} that effectively incorporates the unique characteristics of motion datasets.

Our key insight is that transferring agent behaviors and interactions is crucial for motion planning adaptability. 
Therefore, our method prioritizes preserving and transferring these interactions during the merging process as shown in \cref{fig:viz_intro} (c).
Specifically, our approach consists of two main steps. 
First, during the pre-merging phase, we separately train motion planning models on diverse datasets.
From these models, we construct a pool of metric-specific optimal parameter checkpoints, each maintaining domain-specific interaction patterns.
Second, in the merging phase, the checkpoints are merged into a newly initialized model by trainable weights.
We partition the planning model parameters into key modules that each encapsulate different components of human-robot interaction: human encoder, robot encoder, interaction encoder, and decoder.
In doing so, we divide-and-generalize the heterogeneous nature of human locomotion and interaction.
As a result, our proposed Interaction-Merged Motion Planning (IMMP) can mitigate the issues of domain imbalance and catastrophic forgetting with lower costs.

Our main contributions are as follows:
\begin{itemize}
    \item We propose IMMP, a novel method that merges parameter checkpoints from different domains to enhance adaptation, efficiency, and practicality.
    \item A two-step process extracts diverse motion features and transfers them by identifying key planning parameters, ensuring seamless integration of source domain information into the target domain.
    \item Evaluations on various benchmarks show IMMP’s superior adaptability over \textit{ensemble learning}, \textit{domain adaptation}, \textit{domain generalization}, and other \textit{merging-based approaches}, while integrating into existing frameworks.
\end{itemize}
    
\section{Related Works}
\subsection{Motion Planning}
Motion planning methods mainly utilize grid-based and sampling-based methods.
In grid-based methods, the scene is represented as a grid, and planning is performed using search algorithms such as Dijkstra’s or A* algorithm~\cite{ali2020pathgrid,cheah2018grid,saraydaryan2018navigationgrid,tanzmeister2014efficientgrid,yang2022pathgrid}. 
Sampling-based methods generate random samples and connect them into tree or graph structures to find the optimal plan~\cite{arslan2013usesampling,karaman2011sampling,kavraki1996probabilisticsampling,kuffner2000rrtsampling,wang2020neuralsampling}. 
Recently, deep learning-based approaches have become increasingly popular due to their adaptability to diverse scenarios. 
Earlier deep learning-based planning studies adopted a sequential structure, where perception, prediction, and planning were studied independently~\cite{zhou2022hivt_prediction1, wang2023ganet_prediction2, dong2023sparse_prediction3, cheng2023forecast_prediction4, zhou2023query_prediction5, xu2024adapting_prediction6, yang2024diffusion_planning1, park2023leveraging}. 
Recent trends move towards jointly training forecasters~\cite{kedia2023GameTheoretic, huang2023DIPP, huang2024DTPP, huang2023gameformer} or even integrating all components of the autonomous driving model into the learning process~\cite{zhou2024avatargpt_endtoend1, pan2024vlp_endtoend2, jiang2023vad_endtoend3, jia2023think_endtoend4, weng2024drive_endtoend5, chen2024ppad_endtoend6, hu2023planning_endtoend7, zheng2024genad_endtoend8, wang2024driving_endtoend9}. However, such data-driven approaches are inherently vulnerable to dataset biases.

\subsection{Leveraging Diverse Motion Datasets}
Motion datasets differ in data collection methods, environmental characteristics, dataset size, and diversity. As a result, imitation learning-based models trained for a specific domain often experience performance degradation when deployed in different domains~\cite{gilles2022uncertaintyDomainGeneralization, feng2024unitrajDomainGeneralization, park2024improving, park2024t4p}. To improve robustness in target domains, recent studies have investigated domain generalization at both the dataset and architecture levels. UniTraj~\cite{feng2024unitrajDomainGeneralization} adopts a dataset-level approach by integrating multiple trajectory datasets into a unified dataset, demonstrating that incorporating diverse data enhances generalization. Ye et al.~\cite{ye2023improvingDomainGeneralization} enhance domain generalization by partitioning a single dataset based on vehicle motion characteristics and representing motion in a reference path-based frame. Wang et al.~\cite{wang2023bridgingDomainGeneralization} introduce a module that refines domain-specific velocity and environmental characteristics, while Dong et al.~\cite{dong2024recurrentDomainGeneralization} propose a Stepwise Attention Layer for domain alignment at each timestep. Model architecture-level approaches, such as those in~\cite{ye2023improvingDomainGeneralization, wang2023bridgingDomainGeneralization, dong2024recurrentDomainGeneralization, wang2024forecastDomainGeneralization, xu2022adaptiveDomainGeneralization, qian2024adaptrajDomainGeneralization}, are tailored to specific domains or models, limiting their applicability to general models. In contrast, IMMP operates independently of model architecture, making it adaptable across various planning models.

\subsection{Model Merging}  
Model merging~\cite{yang2024model} integrates parameters from independently trained models to construct a unified model without requiring access to the original training data. This approach can sometimes replace conventional ensemble learning~\cite{arpit2022ensemble, li2022ensemble} or multi-task learning~\cite{kendall2018multi, yu2020gradient, navon2022multi, senushkin2023independent, jeong2024quantifying, liu2024famo} by leveraging only checkpointed parameters. 
Model merging techniques can be broadly categorized into \textit{Pre-Merging} and \textit{During Merging}. Pre-Merging methods~\cite{ortiz2024task, tang2023parameter, jin2024fine, liu2023tangent} fine-tune networks before merging to better align their parameters. Fine-tuning within the tangent space further improves weight disentanglement~\cite{ortiz2024task}, enhancing task arithmetic.
During Merging methods~\cite{ilharco2022editing, demir2024adaptive, wortsman2022model, wortsman2022robust} focus on how parameters are combined. Task arithmetic~\cite{ilharco2022editing}, which adjusts model accuracy by adding or removing task vectors representing parameter differences, plays a crucial role in this process. Ties-Merging~\cite{yadav2023ties} addresses conflicts between models through its trim, elect-sign, and merge process. Despite these advancements, model merging remains largely unexplored in motion planning. Additionally, naive adoption of existing merging approaches has yielded poor results, motivating our proposed merging procedure specifically designed for motion planning.
    
\begin{figure*}[t]
    \vspace{-10pt}
    \centering
    \includegraphics[width=0.99\linewidth]{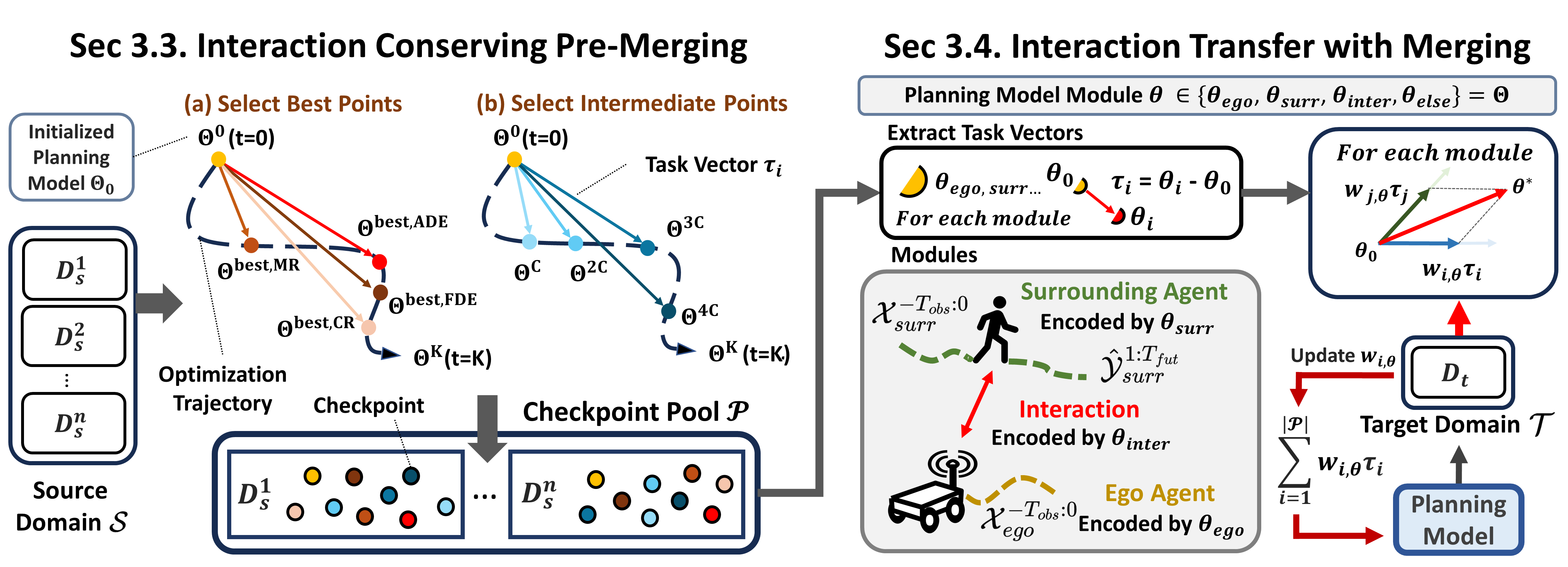}
    \vspace{-8pt}
    \caption{Overview of the proposed IMMP framework. In interaction-conserving pre-merging, parameter checkpoints are selected to preserve the distinct characteristics of different motion planning datasets. In interaction transfer with merging, task vectors are extracted for different modules in the planning model using the collected checkpoint pool $\mathcal{P}$. These task vectors are then merged by learning their linear weights $w$ to enhance adaptation to the target domain.}
    \label{fig:viz_main_method}
    \vspace{-5pt}
\end{figure*}

\section{Method}
\subsection{Problem Definition and Motivation}
\label{sec:problem_def}
 We define samples from the input space $\mathbb{X}$ as $X \in \mathbb{X}$ and from the output space $\mathbb{Y}$ as $Y \in \mathbb{Y}$. 
In the motion planning task, the input space is represented as $X = \{\mathcal{X}_{ego}^{-T_{obs}:0}, \mathcal{X}_{surr}^{-T_{obs}:0}, \hat{\mathcal{Y}}_{surr}^{1:T_{fut}}$\}, 
where $\mathcal{X}_{ego}^{-T_{obs}:0}$ and $\mathcal{X}_{surr}^{-T_{obs}:0}$ are the past trajectory of the ego and surrounding agents, and $\hat{\mathcal{Y}}_{surr}^{1:T_{fut}}$ represents the future predicted trajectories of the surrounding agents. 
We use $\hat{\;}$ to denote prediction.
The output space is $Y = \{\mathcal{Y}_{ego}^{1:T_{fut}}$\}, which represents the future plan of the ego agent.
$T_{obs}$ and $T_{fut}$ respectively represent the observed timesteps of past trajectories and future timesteps of final plan.

The source domains are represented as $\mathcal{S} = \{D_{s}^{1}, \dots, D_{s}^{n}\}$, where $n$ is the number of source domains. The target domains, which we aim to adapt to, are denoted as $\mathcal{T} = \{D_{t}^{1}, \dots, D_{t}^{m}\}$, where $m$ is the number of target domains. Each dataset is expressed as $D^i_s = \{(X^{s,i}_j, Y^{s,i}_j)\}_{j=1}^{l_i}$ for the source domain and $D^i_t = \{(X^{t,i}_j, Y^{t,i}_j)\}_{j=1}^{l_i}$ for the target domain, where $i$ refers to the $i$-th dataset, and $l_i$ is the number of samples within it.
The objective of this work is to leverage $\mathcal{S}$ to enhance the performance of the planning model on $\mathcal{T}$.

Due to the varying characteristics of existing motion planning datasets, domain gaps between source and target domains are more pronounced. Conventional methods such as domain adaptation~\cite{feng2024unitrajDomainGeneralization} and domain generalization~\cite{gilles2022uncertaintyDomainGeneralization} directly utilize source domain samples, incorporating diverse datasets during adaptation. However, these approaches face limitations, as dataset sizes vary significantly, making it difficult to balance their influence during training. Furthermore, this imbalance can also lead to \textit{catastrophic forgetting}, where knowledge from certain domains dominates, causing the model to forget information from other domains. Even with sufficient computational resources, selecting beneficial samples from the source domain $\mathcal{S}$ for adaptation to the target domain $\mathcal{T}$ remains a challenging task. This requires assessing the impact of each sample or domain on performance, which adds complexity to the learning process.

Instead, we propose new adaptation methods that divide the adaptation process into two stages, considering the nature of motion planning: collecting parameter checkpoints (\cref{sec:pre_merge}) and merging them to transfer information across domains (\cref{sec:merge}). The overall IMMP framework is introduced in \cref{fig:viz_main_method}.

\subsection{Model Merging and Task Vector}
Model merging~\cite{yang2024model} is used to integrate the parameters of multiple network backbones, each trained on different tasks, and serves as an alternative to traditional multi-task learning \cite{yu2020gradient, liu2021conflict, liu2023famo, jeong2024quantifying, jeong2025selective} by avoiding joint training. This technique is primarily explored in language-image tasks~\cite{yang2023adamerging, tang2023parameter, ortiz2024task} and image classification~\cite{ilharco2022editing, yadav2023ties}. Notably, Task Arithmetic~\cite{ilharco2022editing} is a pivotal work in this domain, introducing the concept of the Task Vector, which enables simple parameter editing by adding or negating targeted information. The task vector is defined as follows:
\begin{definition}[Task Vector \cite{ilharco2022editing}]
The task vector $\tau_i$ for task $i$ is $\tau_i = \Theta_i - \Theta_0$, where $\Theta_0$ and $\Theta_i$ are the parameters of the pre-trained and fine-tuned models, respectively.
\end{definition}
The model parameters $\Theta$ can then be expressed as a combination of the initial parameters $\Theta_0$ and the task vector set, formulated as follows:
\begin{align}
    \Theta = \Theta_{0} + \lambda \sum_{i=1}^{|\mathcal{P}|} w_i \cdot \tau_i
    \label{eq:theta}
\end{align}
where $w_i$ represents the weight of the task vector $\tau_i$, $\lambda$ is a scaling factor, and $|\mathcal{P}|$ is the number of parameter checkpoints in the pool. The objective of merging is to determine the optimal parameter $\Theta$ for the target domain set $\mathcal{T} = \{D_t^1, \ldots, D_t^m\}$, formulated as:
\begin{align}
    \Theta^* = \arg\min_{\Theta} \sum_{i=1}^{m} \sum_{j=1}^{l_i} \mathcal{L} (\Theta, X^{t,i}_j, Y^{t,i}_j)
\end{align}
Here, $m$ denotes the number of target domains, and $l_i$ represents the number of data samples in the $i$-th target dataset. Since $\Theta_0$ and $\tau_i$ in \cref{eq:theta} are fixed values, merging methods directly optimize $\lambda$ and the weight set $\{w_i\}_{i=1}^{|\mathcal{P}|}$ using train loss.
Initial model parameters $\Theta_0$ can be a model parameter from scratch or one of any fine-tuned models parameters $\{\Theta_i\}_{i=1}^{|\mathcal{P}|}$.
Since the task vector is extracted and linearly weighted during merging, the choice of initial parameters does not drastically affect the results.
Various merging paradigms have been proposed \cite{ilharco2022editing, matena2022merging, yadav2023ties, yang2023adamerging}. However, their direct adaptation to motion planning is either infeasible or insufficient. We will further elaborate on this in the following section, explaining how we address these challenges in motion planning.

\subsection{Interaction-Conserving Pre-Merging}
\label{sec:pre_merge}
There are two major challenges in obtaining usable model checkpoints before the merging process in motion planning. 
First, the motion planning task utilizes multiple metrics, making it challenging to extract the optimal checkpoint.
In classification tasks where merging has traditionally been applied, checkpoint selection is based on a single metric such as classification accuracy, simplifying the process.
However, the evaluation of planning tasks is based on multiple criteria, including effectiveness (ADE), goal success (FDE, Miss Rate), and safety (Collision Rate; CR) where details on the metrics are discussed in Sec.~\ref{sec:metrics_main}.
Since motion planning requires evaluation across multiple criteria, most existing planning models \cite{kedia2023GameTheoretic, huang2023DIPP, huang2024DTPP} employ multiple loss functions, such as trajectory deviation from the ground truth and collision avoidance penalties. 
For simplicity, we denote the sum of all relevant losses as $\mathcal{L}_{total}$.
As $\mathcal{L}_{total}$ consists of multiple loss terms, individual metrics exhibit trade-offs and fluctuations during training.
This complicates the selection of an optimal checkpoint for merging.

Second, prior merging methods rely on pretrained models with large-scale datasets, such as ViT~\cite{vit} pretrained on ImageNet~\cite{deng2009imagenet}. However, this setup differs significantly from motion planning, where models are typically smaller, and no large-scale dataset equivalent to ImageNet exists. As a result, previous works in the vision domain were sufficient to extract only a single checkpoint per domain for merging networks. In contrast, motion planning suffers from data scarcity, necessitating additional methods to expand the checkpoint pool for effective merging.

To address these challenges, we introduce an effective interaction-conserving pre-merging approach, as outlined in \cref{alg:alg1}.
Starting from the parameters $\Theta^{0}$ initialized with $\Theta_{0}$, we iteratively update the model parameters using $\Theta^k = \Theta^{k-1} - \eta \nabla_{\Theta}\mathcal{L}_{total}$ over a total of $K$ iterations with a learning rate $\eta$. 
The superscript of $\Theta$ denotes the iteration, while the subscript refers to the checkpoint pool number.
To overcome the first challenge, which is that various evaluation metrics fluctuate during optimization, we extract parameters corresponding to the best performance for each metric.
Since different metrics emphasize distinct aspects of agent behavior and interactions, the optimal parameters for different metrics reflect different characteristics.
Thus, we collect these best-performing parameters into checkpoint pool $\mathcal{P}$.

As for the second challenge, we leverage the fact that the feature space in intermediate checkpoints during optimization tends to be more generalized across domain shifts, as they are not overly fitted to the source domains. Based on this, we include these intermediate checkpoints at intervals of $C$ from the optimization trajectory in our checkpoint pool.
This process is applied across all source domains.

\begin{algorithm}[t]
\DontPrintSemicolon
\caption{Interaction-Merged Motion Planning}\label{alg:alg1}

\KwData{Source domains $\mathcal{S}=\{D_s^1,\ldots,D_s^n\}$,\quad\quad total iterations $K$, learning rate $\eta$,\quad\quad checkpoint interval $C$, checkpoint pool $\mathcal{P}$, initial model parameters $\Theta^0$}

Initialization: $\mathcal{P} \gets \{\}$ \\
\tcp{\small Interaction-Conserving Pre-Merging}
\For{$D_s^i$ \text{ in } $S$}{
    \For{$k=1\textbf{ to } K$}{
        Update parameters: $\Theta^k \leftarrow \Theta^{k-1}-\eta \nabla_{\Theta} \mathcal{L}_{total}$ \\
        \If{$k \equiv 0 \pmod{C}$}
            {
                $\mathcal{P} \leftarrow \mathcal{P} \cup \{\Theta^k\}$
            }
        \For{$m \in \{ADE, FDE, CR, MR\}$}{
            \If{$m^k$ \text{is better than} $m^{best}$}
            {
                $\Theta^{best,m} \leftarrow \Theta^k$ \\
                $m^{best} \leftarrow m^k$
            }
        }
    }
}
\For{$m \in \{ADE, FDE, CR, MR\}$}{
    $\mathcal{P} \leftarrow \mathcal{P} \cup \{\Theta^{best,m}\}$}

\tcp{\small Interaction Transfer with Merging}
Let $\mathcal{P} = \{\Theta_1, \ldots, \Theta_{|\mathcal{P}|}\}$ \\
\For{$\theta \in \{\theta_{ego}, \theta_{surr}, \theta_{inter}, \theta_{else}\}$}{
    \For{$i=1$ \textbf{ to } $|\mathcal{P}|$}{
        Compute task vector: $\tau_i = \theta_i - \theta_0$
    }
    Merge parameters: $\theta^* = \theta_{0} + \sum_{i=1}^{|\mathcal{P}|} w_{i,\theta} \cdot \tau_i$
}
Update all $\{w_{i, \theta}\}_{i=1}^{|\mathcal{P}|}$ using the train set and $\mathcal{L}_{total}$.
\end{algorithm}

\subsection{Interaction Transfer with Merging}
\label{sec:merge}
Our goal is to effectively transfer appropriate planning-related information from the source to the target domains.
To this end, our idea starts with the observation that each source dataset contains unique ego and surrounding agents past trajectories and interaction information.
Therefore, we separate and respectively merge the modules for each distinct features, enabling a better adaptation on the unique characteristics of the target domain.
This approach differs from traditional merging techniques, such as model-level merging and parameter-wise merging. As shown in Tab.~\ref{tab:granularity}, extracting key features individually within the source domain is more efficient than traditional merging methods. This demonstrates that, for domain-robust motion planning, it is crucial to properly extract key modules during the merging process.
To ensure the general applicability of the IMMP methodology to various planning models, we categorize the modules that are essential in planning networks.

Conventional planning models employ hierarchical structures to encode agent behavior and their interactions. For instance, the trajectory information of the ego agent and surrounding agents is embedded using the modules $\phi_{ego}$ and $\phi_{surr}$, typically implemented with LSTMs.
\begin{gather}
    h_{ego} = \phi_{ego}(\mathcal{X}_{ego}^{-T_{obs}:0}; \theta_{ego}) \\
    h_{surr} = \phi_{surr}(\mathcal{X}_{surr}^{-T_{obs}:0}, \hat{\mathcal{Y}}_{surr}^{1:T_{fut}}; \theta_{surr})
\end{gather}
where $\theta_{ego}$ and $\theta_{surr}$ are the parameters of each module. 
Interaction information is embedded based on $h_{ego}$ and $h_{surr}$ using the $\psi_{inter}$ module, which is typically implemented with a Transformer.
\begin{gather}
    h_{inter} = \psi_{inter}(h_{surr}, h_{ego}; \theta_{inter})
\end{gather}
where $\theta_{inter}$ is the parameters of $\psi_{inter}$.
Finally, the planning $\hat{\mathcal{Y}}_{ego}^{1:T_{fut}}$ is generated based on the decoder using $h_{inter}$.
The detailed model structure of the planning models used in the experiments is in suppl B.

Based on the key modules of the planning model that learn features with significant differences between the source domains, we group the parameters $\theta_{ego}, \theta_{surr},$ and $\theta_{inter}$ while classifying the remaining decoder and other layers as $\theta_{else}$.
Thus, the complete set of model parameters is defined as $\Theta = \{\theta_{ego}, \theta_{surr}, \theta_{inter}, \theta_{else}\}$.  
We perform merging separately for each parameter group, assigning distinct merging weights $\{w_{i, \theta}\}_{i=1}^{|\mathcal{P}|}$ to each $\theta$ in $\{\theta_{ego}, \theta_{surr}, \theta_{inter}, \theta_{else}\}$. This enables interaction-level information merging for effective adaptation in planning models while preserving the feature hierarchy. The overall procedure is outlined in the interaction transfer with merging step in \cref{alg:alg1}.

Relying solely on model parameters during adaptation provides two key advantages. First, it eliminates the need to access source domains after extracting the necessary parameter checkpoints, reducing the adaptation cost for the target domain with a pre-constructed checkpoint pool.
Second, it effectively mitigates domain imbalance and catastrophic forgetting, which arise from imbalanced dataset compositions that affect previous approaches.

\section{Experiments}
\subsection{Experimental Setup}
\noindent\textbf{Datasets \& Backbone.} In our experiments, we utilize three categories of datasets: ETH-UCY~\cite{pellegrini2009ETH,lerner2007UCY} (Human-Human Interaction dataset), CrowdNav~\cite{chen2019CrowdNav} (RL algorithm-based Robot dataset), THOR~\cite{rudenko2020THOR} and SIT~\cite{bae2023SIT} (Human-Robot Interaction dataset). 
For ETH-UCY, we leverage the unique characteristics of each sub-dataset while maintaining the overall dataset size by alternately selecting 4 out of 5 scenes to compose source datasets. The excluded scene is denoted as ``w/o" in our notation.
To extract the key characteristics of each dataset appropriately, we apply a preprocessing step to standardize all data into a uniform format. Specifically, we extract ego-centric x, y coordinates from all datasets and sample the data at 2.5 FPS. Additionally, the observed past trajectories consist of 8 timesteps, while the predicted trajectories of surrounding agents and the planned trajectory of the ego agent are set to 12 timesteps. 
In our experiments, we vary the target domain across different datasets, using SIT and THOR as target domains. 
Considering practical applicability, THOR and SIT are selected as target datasets, as they are relatively small so require more information transfer from other datasets.
The datasets not included in the target domain are treated as the source domain.
We evaluate the performance of our methodology against three planning baseline models:  \textbf{GameTheoretic}~\cite{kedia2023GameTheoretic},  \textbf{DIPP}~\cite{huang2023DIPP}, and  \textbf{DTPP}~\cite{huang2024DTPP}. 
Detailed descriptions of each data collection process and the backbone planning models are provided in suppl A and B.

\vspace{2pt}
\noindent\textbf{Metrics.} \label{sec:metrics_main} The primary evaluation criteria for autonomous robot driving include Effectiveness, Safety, and Goal Success. Average Displacement Error (ADE) is used as the metric for Effectiveness, Collision Rate (CR) for Safety, and Miss Rate (MR) along with Final Displacement Error (FDE) for Goal Success. Additionally, the final inference cost, such as FLOPs or the number of model parameters, is measured as Cost. Since all methods use the same network architecture, Cost is reported as a relative multiplier with respect to the baseline model. More details on the evaluation metrics are provided in suppl C.

\vspace{2pt}
\noindent\textbf{Baselines.} Our approach is not model-specific and can be applied orthogonally with other methods, so we use non-model-specific methods as baselines and compare them with our IMMP.
(1) \textbf{Target Only} trains a model exclusively on the target dataset. (2) \textbf{Domain Generalization} \cite{gilles2022uncertaintyDomainGeneralization} pretrains the model on all datasets except the target dataset. (3) \textbf{Domain Adaptation} \cite{feng2024unitrajDomainGeneralization} pretrains the model on all datasets except the target dataset and then finetunes it on the target dataset. (4) \textbf{Ensemble} \cite{arpit2022ensemble, li2022ensemble} uses multiple models, each on a different source dataset, and combines their trajectories for the final planning. Specifically, we incorporate two ensemble strategies for comparison: winner-takes-all (WTA) and averaging (AVG). Additionally, we incorporate model merging approaches such as (5) \textbf{Averaging} \cite{wortsman2022model, choshen2022fusing}, which computes the parameter-wise mean of all individual models, defined as $\theta_{merge} = \sum_{i=1}^{n} \theta_i/n$. (6) \textbf{Task Arithmetic} \cite{ilharco2022editing} sums all task vectors and scales them to produce a merged model, formulated as $\theta_{merge} = \theta_0 + \lambda \cdot \sum_{i=1}^{n} \tau_i$. (7) \textbf{Ties Merging} \cite{yadav2023ties} performs the merge in three steps: trimming task vectors with minimal change during training, resolving sign conflicts, and merging only parameters that align with the final signed agreed on.

\subsection{Experimental Results}
\noindent\textbf{Merging Interaction for Better Initialization.} IMMP not only improves generalization for target domains but also provides better initialization for fine-tuning to target domains. \Cref{tab:main} presents IMMP's performance after directly merging parameter checkpoints, while IMMP+Finetune shows the results of fine-tuning the merged model parameters used as initialization. Notably, IMMP+Finetune outperforms both Domain-Specific and Domain Adaptation scenarios. This result supports that IMMP offers a strong initialization point for fine-tuning.

\vspace{2pt}  
\noindent\textbf{Domain Gap and Generalization.}  
\Cref{tab:main} presents a comparison of IMMP with baseline methods. Domain generalization for motion planning, which incorporates multiple source datasets~\cite{feng2024unitrajDomainGeneralization}, shows significantly lower performance. Interestingly, training solely on the target domain outperforms domain generalization~\cite{gilles2022uncertaintyDomainGeneralization}, highlighting the substantial domain disparity between source and target datasets. Additionally, domain adaptation does not always achieve better performance than the target-only approach, further supporting this observation.  
Ensemble-WTA significantly outperforms Ensemble-AVG, suggesting that only certain source domains effectively contribute to target domain performance. These results indicate potential limitations in previous methods, such as domain imbalance or catastrophic forgetting. In contrast, IMMP leverages parameters trained across different domains, reducing susceptibility to these issues.

\begin{table*}
    \caption{Qualitative comparison of the proposed IMMP with baseline methods. Each baseline is implemented across various planning models. The methods are categorized into conventional approaches for generalization in motion planning, adaptations of other model merging methods in our setting, and the proposed IMMP.}
    \vspace{-5pt}
    \renewcommand{\arraystretch}{1.2}
    \centering
    \resizebox{1\linewidth}{!}{
    \begin{tabular}{c|l|cccc|cccc|c}
    \hline
    \rowcolor{gray!30}
    \multicolumn{2}{c|}{Target Dataset} & \multicolumn{4}{c|}{SIT~\cite{bae2023SIT}} & \multicolumn{4}{c|}{THOR~\cite{rudenko2020THOR}} & \\
    \cline{1-10} \rowcolor{gray!30}
     & & Effectiveness & Safety & \multicolumn{2}{c|}{\multirow{1}{*}{Goal Success}} & Effectiveness & Safety & \multicolumn{2}{c|}{\multirow{1}{*}{Goal Success}} & Cost \\ \rowcolor{gray!30}
    \multicolumn{1}{c|}{\multirow{-2}{*}{Model}}& \multicolumn{1}{c|}{\multirow{-2}{*}{Method}} & ADE $\downarrow$ & Col. Rate $\downarrow$ & FDE $\downarrow$ & Miss Rate $\downarrow$ & ADE $\downarrow$ & Col. Rate $\downarrow$ & FDE $\downarrow$ & Miss Rate $\downarrow$ & \\
    \hline
    \multirow{10}{*}{GameTheoretic \cite{kedia2023GameTheoretic}}
     & Domain Generalization \cite{gilles2022uncertaintyDomainGeneralization}     &0.8338&9.87E-04&1.8594&0.9355      &0.3804&1.21E-03&0.8705&0.6957&$\times 1$ \\
     & Domain Adaptation \cite{feng2024unitrajDomainGeneralization}                                                         &0.4388&1.26E-03&1.0611&0.7201      &0.1133&2.90E-04&0.2516&0.1268&$\times 1$ \\
     & Target Only         &0.4343&3.41E-04&0.9014&0.6272      &0.1003&2.64E-04&0.2153&0.0929&$\times 1$ \\
     & Ensemble-WTA \cite{arpit2022ensemble}                                &0.3695&5.75E-05&0.8283&0.6185      &0.2112&4.03E-04&0.4082&0.2996&$\times 7$ \\
     & Ensemble-AVG \cite{arpit2022ensemble}                                &0.5415&1.89E-04&1.1617&0.8159      &0.3142&4.86E-04&0.6181&0.6027&$\times 7$ \\\cline{2-11}
     & Averaging \cite{wortsman2022model, choshen2022fusing}                &0.6726&2.34E-04&1.4340&0.9611      &0.2742&5.17E-04&0.5624&0.5711&$\times 1$ \\
     & Task Arithmetic \cite{ilharco2022editing}                            &0.4132&1.37E-04&0.8936&0.7364      &0.2679&3.87E-04&0.5651&0.4812&$\times 1$ \\
     & Ties Merging \cite{yadav2023ties}                                    &1.1876&5.53E-04&2.2440&0.9872      &0.5253&2.49E-03&0.8212&0.6201&$\times 1$ \\\cline{2-11}
     & IMMP \cellcolor{blue!10}                                             &\cellcolor{blue!10}0.3380&\cellcolor{blue!10}5.12E-05&\cellcolor{blue!10}0.7626&\cellcolor{blue!10}0.6446      &\cellcolor{blue!10}0.1165&\cellcolor{blue!10}3.48E-04&\cellcolor{blue!10}0.2562&\cellcolor{blue!10}0.1330&\cellcolor{blue!10}$\times 1$ \\
     & IMMP + Finetune  \cellcolor{blue!10}                                                    &\cellcolor{blue!10}\textbf{0.3157}&\cellcolor{blue!10}\textbf{4.28E-05}&\cellcolor{blue!10}\textbf{0.7300}&\cellcolor{blue!10}\textbf{0.5934}      &\cellcolor{blue!10}\textbf{0.0975}&\cellcolor{blue!10}\textbf{2.56E-04}&\cellcolor{blue!10}\textbf{0.2108}&\cellcolor{blue!10}\textbf{0.0912}&\cellcolor{blue!10}$\times 1$ \\
    \hline
    \multirow{10}{*}{DTPP \cite{huang2024DTPP}} 
     & Domain Generalization \cite{gilles2022uncertaintyDomainGeneralization}     &1.0116 &1.18E-03 &2.1255 &0.9350 &0.3446 &1.42E-03 &0.6873 &0.5948 &$\times 1$ \\
     & Domain Adaptation \cite{feng2024unitrajDomainGeneralization}                                                          &0.4640 &2.45E-04 &1.0116 &0.7573 &\textbf{0.1460} &2.50E-04 &\textbf{0.2291} &\textbf{0.0772} &$\times 1$ \\
     & Target Only         &0.4832 &5.73E-04 &0.9839 &0.6852 &0.1528 &2.57E-04 &0.2394 &0.0910 &$\times 1$ \\
     & Ensemble-WTA \cite{arpit2022ensemble}                                &0.6125 &\textbf{2.01E-04} &1.2067 &0.8217 &0.2241 &5.15E-04 &0.4591 &0.4127 &$\times 7$ \\
     & Ensemble-AVG \cite{arpit2022ensemble}                                &0.7460 &3.36E-04 &1.5439 &0.9175 &0.2145 &3.41E-04 &0.4095 &0.3302 &$\times 7$ \\\cline{2-11}
     & Averaging \cite{wortsman2022model, choshen2022fusing}                &0.6303 &2.79E-04 &1.2699 &0.8641 &0.3075 &6.75E-04 &0.5807 &0.4901 &$\times 1$ \\
     & Task Arithmetic \cite{ilharco2022editing}                            &0.6062 &3.73E-04 &1.2138 &0.8542 &0.2885 &4.18E-04 &0.5057 &0.4693 &$\times 1$ \\
     & Ties Merging \cite{yadav2023ties}                                    &1.0741 &2.25E-03 &2.0873 &0.9384 &0.5141 &2.15E-03 &0.6764 &0.5514 &$\times 1$ \\\cline{2-11}
     & IMMP \cellcolor{blue!10}                                                                &0.4388\cellcolor{blue!10} &2.58E-04\cellcolor{blue!10} &0.9067\cellcolor{blue!10} &\textbf{0.6220}\cellcolor{blue!10} &0.1958\cellcolor{blue!10} &2.46E-04\cellcolor{blue!10} &0.3210\cellcolor{blue!10} &0.1558\cellcolor{blue!10} &$\times 1$\cellcolor{blue!10} \\
     & IMMP + Finetune   \cellcolor{blue!10}                                                   &\textbf{0.3793}\cellcolor{blue!10} &4.22E-04\cellcolor{blue!10} &\textbf{0.8336}\cellcolor{blue!10} &0.6336\cellcolor{blue!10} &0.1470\cellcolor{blue!10} &\textbf{2.17E-04}\cellcolor{blue!10} &0.2421\cellcolor{blue!10} &0.0905\cellcolor{blue!10} &$\times 1$\cellcolor{blue!10} \\
    \hline
    \multirow{10}{*}{DIPP \cite{huang2023DIPP}}
     & Domain Generalization \cite{gilles2022uncertaintyDomainGeneralization}     &1.3268&1.09E-03&2.7111&0.9756      &0.3398 &1.27E-03 &0.6983 &0.6163 &$\times 1$ \\
     & Domain Adaptation \cite{feng2024unitrajDomainGeneralization}                                                          &0.4697&3.93E-04&1.0331&0.7886      &0.2660 &9.68E-04 &0.5272 &0.4773 &$\times 1$ \\
     & Target Only         &0.5671&6.68E-04&0.9801&0.7253      &0.1771 &2.17E-04 &0.2713 &0.1156 &$\times 1$ \\
     & Ensemble-WTA \cite{arpit2022ensemble}                                &0.7784&4.36E-04&1.5916&0.8275      &0.2604 &7.00E-04 &0.4874 &0.4435 &$\times 7$ \\
     & Ensemble-AVG \cite{arpit2022ensemble}                                &0.9737&5.79E-04&1.9442&0.9750      &0.2562 &5.46E-04 &0.5134 &0.4754 &$\times 7$ \\\cline{2-11}
     & Averaging \cite{wortsman2022model, choshen2022fusing}                &1.4383&1.74E-03&2.7763&0.9820      &0.4329 &1.47E-03 &0.7534 &0.6710 &$\times 1$ \\
     & Task Arithmetic \cite{ilharco2022editing}                            &1.2614&1.66E-03&2.4944&0.9564      &0.4196 &1.39E-03 &0.7120 &0.6915 &$\times 1$ \\
     & Ties Merging \cite{yadav2023ties}                                    &1.6028&1.85E-03&3.0906&1.0000      &0.5141&2.15E-03&0.6764&0.5515&$\times 1$ \\\cline{2-11}
     & IMMP \cellcolor{blue!10}                                                                &\cellcolor{blue!10}0.5112&\cellcolor{blue!10}\textbf{1.60E-04}&\cellcolor{blue!10}0.9358&\cellcolor{blue!10}0.7944      &\cellcolor{blue!10}0.2335 &\cellcolor{blue!10}5.87E-04 &\cellcolor{blue!10}0.3219 &\cellcolor{blue!10}0.1842 &\cellcolor{blue!10}$\times 1$ \\
     & IMMP + Finetune \cellcolor{blue!10}                                                     &\cellcolor{blue!10}\textbf{0.4096}&\cellcolor{blue!10}5.27E-04&\cellcolor{blue!10}\textbf{0.8915}&\cellcolor{blue!10}\textbf{0.6789}      &\cellcolor{blue!10}\textbf{0.1204} &\cellcolor{blue!10}\textbf{1.94E-04} &\cellcolor{blue!10}\textbf{0.2236} &\cellcolor{blue!10}\textbf{0.0769} &\cellcolor{blue!10}$\times 1$ \\
    \hline
  \end{tabular}
  }
  \label{tab:main}
  \vspace{-5pt}
\end{table*}

\begin{table}[t]
\caption{Impact of merging granularity in IMMP for SIT Datasets with GameTheoretic. The granularity levels include Model-level, Parameter-level, and Interaction-level.}
\vspace{-5pt}
\renewcommand{\arraystretch}{1.1}
\resizebox{0.95\linewidth}{!}{
\begin{tabular}{l|cccc}
\hline
\rowcolor{gray!30}
\multicolumn{1}{c}{} & \multicolumn{1}{|c}{Effectiveness} & \multicolumn{1}{c}{Safety} & \multicolumn{2}{c}{\multirow{1}{*}{Goal Success}} \\ \rowcolor{gray!30}
\multicolumn{1}{c}{\multirow{-2}{*}{Granularity}}                             & \multicolumn{1}{|c}{ADE $\downarrow$}      & \multicolumn{1}{c}{Col. Rate $\downarrow$}     & \multicolumn{1}{c}{FDE $\downarrow$}          & \multicolumn{1}{c}{Miss Rate $\downarrow$}    \\ \hline
Model                                            &0.3687                                   &7.15E-05             
            &0.8365                              &0.8002 \\
Parameter                                        &0.3798                                   &9.16E-05                                  &0.7754                       &0.7433 \\
Interaction                                      &\textbf{0.3380}                          &\textbf{5.12E-05}                         &\textbf{0.7626}              &\textbf{0.6446} \\
\bottomrule           
\end{tabular}}
\label{tab:granularity}
\vspace{-10pt}
\end{table}

\vspace{2pt}
\noindent\textbf{Comparison to Merging Approaches.} 
\Cref{tab:main} shows that the naive adoption of existing model merging techniques leads to poor performance in motion planning. 
Previous methods fail to consider the feature hierarchy specific to motion planning, where features can be encoded at the agent behavior and interaction levels. As a result, motion planning models exhibit significant disparities after training, making them difficult to merge using conventional methods. Specifically, Averaging and Task Arithmetic can disrupt the hierarchy of features, while Ties Merging tends to discard too much critical information during the trimming process when resolving conflicts.  
In contrast, IMMP efficiently transfers information from source datasets to the target domain, achieving superior performance.

\begin{figure*}[t]
    \centering
    \vspace{-10pt}
    \includegraphics[width=0.95\linewidth]{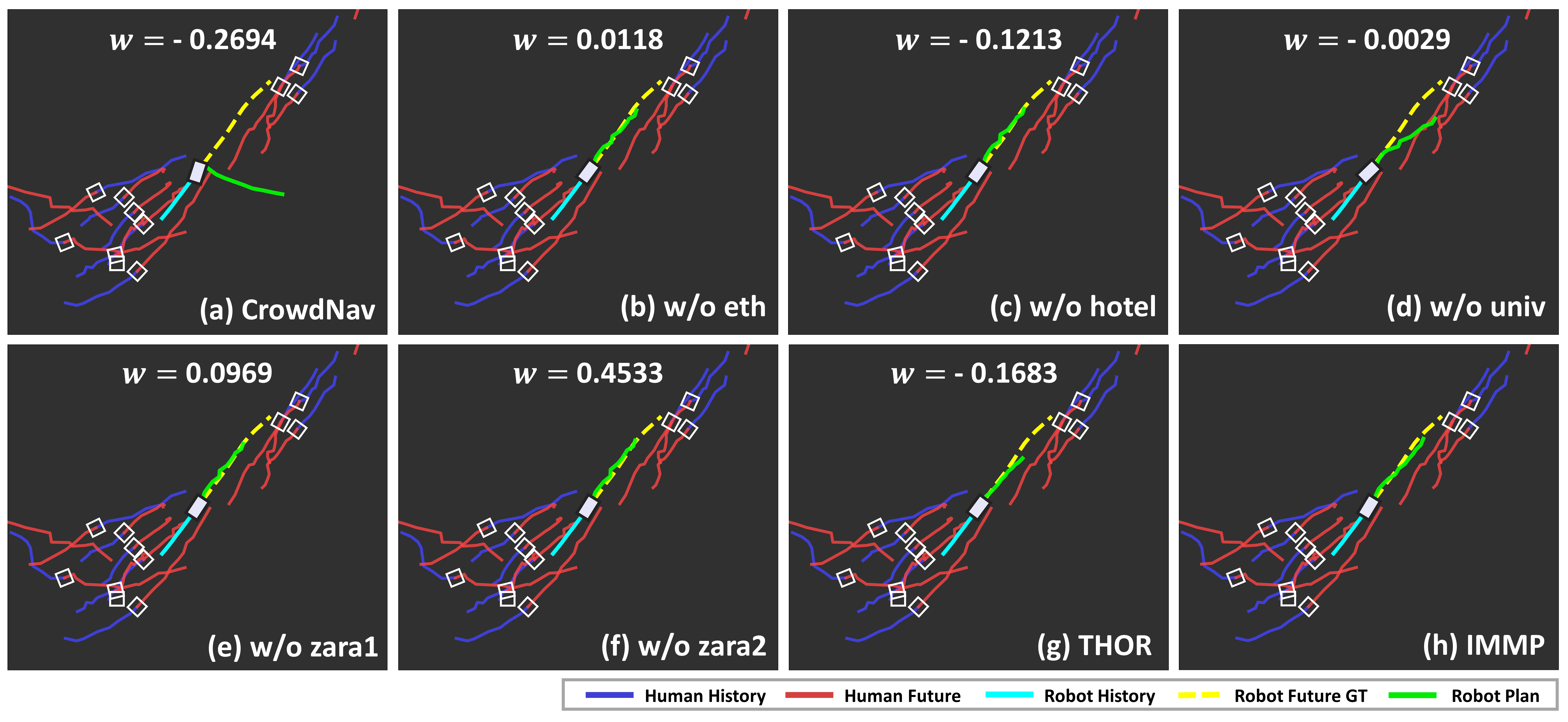}
    \vspace{-8pt}
    \caption{Qualitative results on the SIT dataset with GameTheoretic planning model. (a) to (g) represent the inference results of SIT using models trained on individual dataset from the source domains. The values above, denoted by $w$, indicate the average of contributions of task vectors per epoch and per module for each source domain in the IMMP planner.}
    \label{fig:viz_qual}
    \vspace{-7pt}
\end{figure*}

\vspace{2pt}
\noindent\textbf{Qualitative Results.}
In \cref{fig:viz_qual}, we qualitatively analyze how task vectors extracted from each dataset are utilized. 
Figure.~\ref{fig:viz_qual} (a) to (g) illustrate the inference results in the target domain SIT using GameTheoretic models trained on individual source domains, while \cref{fig:viz_qual} (h) presents the inference result of the planner after interaction transfer merging in the target domain.
The weights above the inference results represent the actual merging weights applied by the proposed IMMP. Notably, there is a meaningful relationship between the merging weights and the performance of individual domains. As shown in ~\cref{fig:viz_qual} (a) and (g), source datasets with poor inference performance in the target domain are assigned lower weights, whereas well-performing domain, such as~\cref{fig:viz_qual} (f), is assigned higher weight and utilized more extensively. This demonstrates that our approach effectively prioritizes source domain datasets that are most beneficial for the target domain.

\subsection{Ablation Study}
\noindent\textbf{Merging Granularity.} Table~\ref{tab:granularity} evaluates the impact of merging granularity in SIT datasets using GameTheoretic. We consider three levels: `Model-level,' where identical merging weights apply to all parameters; `Parameter-level,' where merging is parameter-wise; and `Interaction-level,' which merges components related to past ego-trajectory, surrounding agents, and their interactions. Results show that Interaction-level merging achieves the best performance across all metrics. Model-level merging fails to adapt interactions and agent behaviors from source to target domains, as full-model transfer lacks flexibility for domain-specific interactions. Parameter-level merging allows finer adaptation but disrupts the feature hierarchy of trajectory encoding and interaction modeling, introducing instability. While it offers slight improvements over Model-level merging, it does not outperform Interaction-level merging in most cases. These findings underscore the need for proper merging granularity to retain interaction structures and improve planning model adaptation in the target domain.

\begin{table}[t]
\vspace{2pt}
\caption{Ablation study on checkpoint types and pool size for merging. IMMP planning performance with best checkpoints (across all metrics) and intermediate-epoch for planning (P) / forecaster (F) models; fine-tuning results included.}
\vspace{-5pt}
\renewcommand{\arraystretch}{1.1}
\resizebox{1\linewidth}{!}{
\begin{tabular}{cc|c|cccc}
\hline
\rowcolor{gray!30}
\multicolumn{2}{c|}{Selection Method} &  & \multicolumn{1}{c}{Effectiveness}& \multicolumn{1}{c}{Safety} & \multicolumn{2}{c}{\multirow{1}{*}{Goal Success}} \\ \cline{1-2} \rowcolor{gray!30}
 All Metric &  \multicolumn{1}{c|}{Epoch} & \multicolumn{1}{c|}{\multirow{-2}{*}{Finetune}} & \multicolumn{1}{c}{ADE $\downarrow$}        & \multicolumn{1}{c}{Col. Rate $\downarrow$}   & \multicolumn{1}{c}{FDE $\downarrow$}      & \multicolumn{1}{c}{Miss Rate $\downarrow$}    \\ \hline
                   &                  &             &0.3646   &8.12E-05 &0.8063   &0.6516 \\
    \checkmark     &                  &             &0.3543   &6.44E-05 &0.7730   &\textbf{0.6196} \\
    \checkmark     &\checkmark (P)    &             &0.3390   &\textbf{5.12E-05} &0.7639   &0.6400 \\
    \checkmark     &\checkmark (F,P)  &             &\textbf{0.3380}   &\textbf{5.12E-05} &\textbf{0.7626}   &0.6446 \\ \hline
                   &                  &\checkmark   &0.3203   &5.40E-05 &0.7365	 &\textbf{0.5743} \\
    \checkmark     &                  &\checkmark   &0.3186   &5.09E-05 &0.7358   &0.5981 \\
    \checkmark     &\checkmark (P)    &\checkmark   &0.3176   &4.31E-05 &0.7399   &0.5906 \\
    \checkmark     &\checkmark (F,P)  &\checkmark   &\textbf{0.3157}   &\textbf{4.28E-05} &\textbf{0.7300}   &0.5934 \\
    
\bottomrule           
\end{tabular}}
\label{tab:checkpoint_pool}
\vspace{-8pt}
\end{table}

\vspace{2pt}
\noindent\textbf{Impact of Checkpoint Types and Pool Size.} Table~\ref{tab:checkpoint_pool} evaluates how the composition of the checkpoint pool $\mathcal{P}$ affects planning performance using GameTheoretic. `All Metric' selects checkpoints based on the best performance across all evaluation metrics, while its absence defaults to ADE-based selection. `Epoch' includes intermediate checkpoints along the optimization trajectory. Given the separate forecaster and planner networks in GameTheoretic, we independently select their checkpoints. For instance, (P) includes only intermediate checkpoints of the planning network, while (F,P) incorporates both. Fine-tuning on the target domain and increasing the checkpoint pool size improve performance after both merging and fine-tuning. Notably, selecting checkpoints using multiple metrics significantly enhances their corresponding evaluation scores, as seen in the superior performance of `All Metric' over ADE-only selection. Intermediate checkpoints further benefit adaptation, preventing overfitting to the source domain. However, these results do not necessarily imply that the pre-merging phase is essential, as IMMP still achieves promising performance compared to previous approaches when using only checkpoints selected based on `ADE'.

\section{Conclusion}
In this paper, we present the Interaction-Merged Motion Planning (IMMP), which effectively leverages diverse motion planning datasets. By utilizing parameter checkpoints trained on different domains, IMMP addresses domain imbalance and catastrophic forgetting that hinder conventional adaptation approaches. The proposed two-step process—interaction-conserving pre-merging and interaction-level merging—preserves agent behavior and interactions, resulting in a more adaptable planning model. Experimental results demonstrate that IMMP outperforms traditional approaches, achieving superior performance.

\section*{Acknowledgment}
This work was supported by the National Research Foundation of Korea(NRF) grant funded by the Korea government(MSIT) (NRF2022R1A2B5B03002636), and by the Institute of Information \& communications Technology Planning \& Evaluation (IITP) grant funded by the Korea government(MSIT) (No. RS-2024-00457882, AI Research Hub Project).

{
    \small
    \bibliographystyle{ieeenat_fullname}
    \bibliography{main}
}

\clearpage
\appendix
\clearpage
\maketitlesupplementary

\section{Datasets}
\label{sup:Datasets}
\begin{enumerate}
    \item \textbf{Human-Human Interaction Dataset: ETH-UCY}

    The ETH-UCY dataset~\cite{pellegrini2009ETH,lerner2007UCY} consists of five sub-datasets: ETH, Hotel, Univ, Zara1, and Zara2, each with distinct pedestrian densities and scene characteristics. 
    We alternately select 4 out of 5 scenes to form the training and validation datasets, and train a separate model for each configuration. 
    Additionally, since this dataset does not include a designated robot, we construct the dataset by alternately assuming each human agent in the scene as the robot.

    \item \textbf{RL algorithm-based Robot Dataset: CrowdNav}

    The CrowdNav dataset~\cite{chen2019CrowdNav} is a simulation-based dataset designed to enable collision-free navigation in crowded environments. To model human interactions, the dataset first generates human movements by employing the ORCA algorithm, allowing agents to reach their destinations while avoiding collisions. Subsequently, reinforcement learning is used to generate the robot’s trajectory, ensuring it navigates without colliding with humans, thereby forming the complete dataset. Following the dataset composition approach from the previous study~\cite{kedia2023GameTheoretic}, we split the dataset into a 50:50 ratio for the training and validation sets.

    \item \textbf{Human-Robot Interaction Dataset: THOR and SIT}

    The THOR dataset~\cite{rudenko2020THOR} is collected in an indoor environment with real humans and a robot. The data was gathered in an indoor space measuring 8.4 × 18.8 m, with various fixed obstacles placed throughout. In this setting, real humans navigate toward one of five designated destinations while avoiding the moving robot. Meanwhile, the robot follows a predetermined path to patrol the indoor space, regardless of nearby human presence and without considering human interaction. Unlike the CrowdNav dataset, where the robot takes human interaction into account, the THOR dataset captures scenarios where humans adjust their movement in response to the robot.

    The SIT dataset~\cite{bae2023SIT} contains indoor and outdoor scenes where both robots and humans move while considering their interactions. Data was collected from a total of 10 different scenes, each with varying crowd densities. Since the test set does not provide ground truth position information for surrounding agents, we conduct evaluations using the validation set. 
\end{enumerate}


\begin{figure*}[t]
    \centering
    \vspace{-5pt}
    \includegraphics[width=0.99\linewidth]{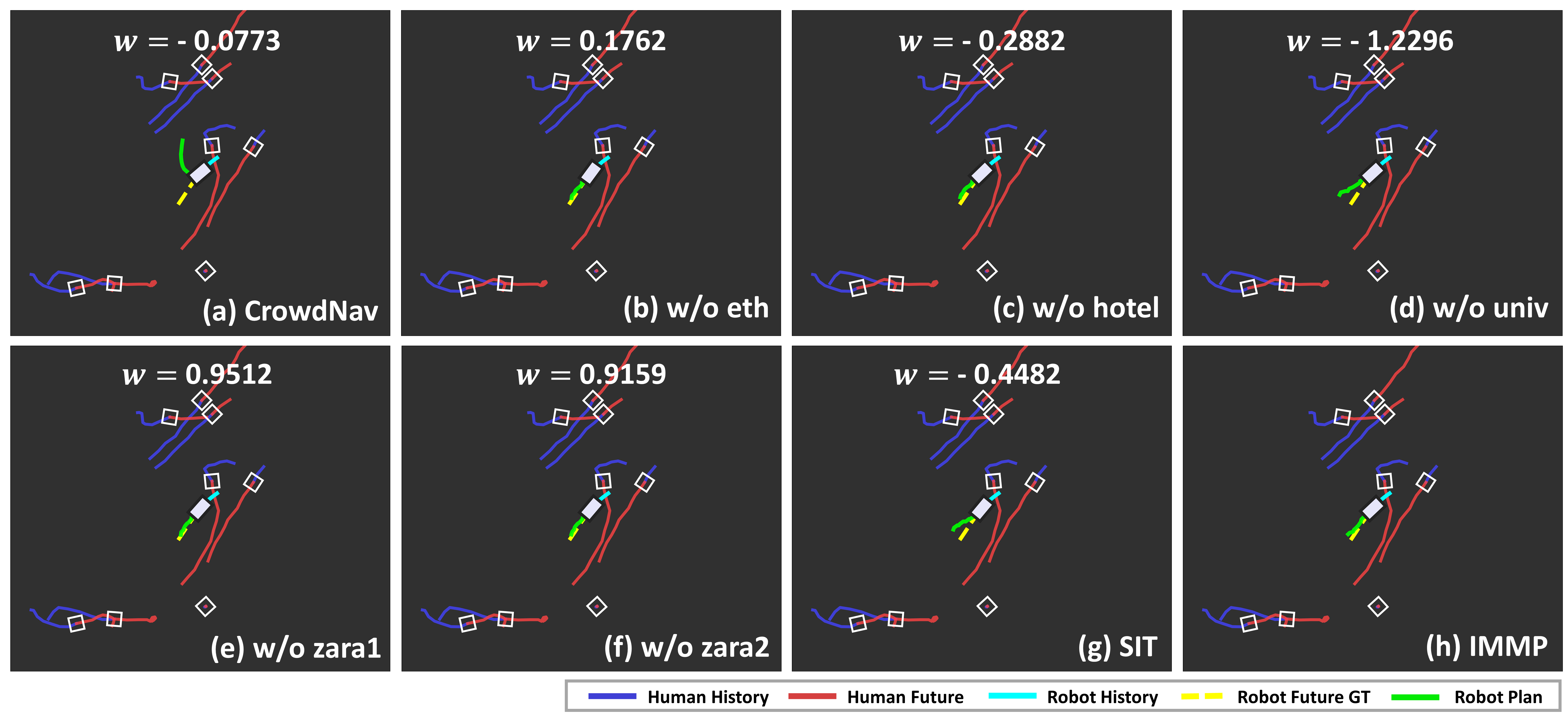}
    \vspace{-5pt}
    \caption{Qualitative results on the THOR dataset with GameTheoretic planning model. From (a) to (g) represent the inference results for each model trained on source domains. The values $w$ indicate the average of weights of task vectors for each source domain in the IMMP planner.}
    \label{fig:viz_qual_THOR_supple}
    \vspace{-5pt}
\end{figure*}

\section{Backbone Models}
\label{sup:BackboneModels}
We conduct experiments using three different backbone models. 
Unlike the GameTheoretic model, DIPP and DTPP are designed for vehicle datasets, requiring some modifications to adapt them to the given robot dataset.
(1) \textbf{GameTheoretic}~\cite{kedia2023GameTheoretic} trains the forecaster to generate risky forecasts for the ego agent, encouraging the planner to produce safer plans. To extract a valid task vector from a consistent initial state across different datasets, we omit the step of training the forecaster to be risky. Instead, we ensure that the planner considers collisions during its training process. (2) \textbf{DIPP}~\cite{huang2023DIPP} integrates the prediction module with the planning module, jointly training the prediction model to improve planning performance. Since the planner is trained after the predictor, the task vector is extracted from epochs after the planner's training begins. Additionally, weights are extracted at once from the unified module combining the planner and predictor.
(3) \textbf{DTPP}~\cite{huang2024DTPP} constructs a tree-structured planner and selects the optimal plan based on cost, achieving higher performance than single-step planning methods that directly generate plans. In this study, we use the model as is, incorporating it as a baseline.

The GameTheoretic uses both the past trajectory of surrounding agents, $\mathcal{X}_{surr}^{-T_{obs}:0}$, and the future predicted trajectory $\hat{\mathcal{Y}}_{surr}^{1:T_{fut}}$ generated by the forecaster to embed surrounding agents' feature $h_{surr}$. 
This feature is then used as the input to the interaction layer $\psi_{inter}$.
In contrast, for DIPP and DTPP, $h_{surr}$ is constructed solely based on the past trajectory of surrounding agents, $\mathcal{X}_{surr}^{-T_{obs}:0}$, and used as the input to the interaction layer. Instead, the loss for both the plan and prediction generated by the independent decoder from the transformer is provided to simultaneously train the forecaster and planner in a meaningful way.

\section{Evaluation Metrics}
\label{sup:eval}
\begin{enumerate}
\item \textbf{Average Displacement Error (ADE)}

ADE is a metric for evaluating effectiveness by assessing how similar the generated ego agent's future plan is to the dataset's ground truth trajectory. ADE computes the L2 distance between every time step of the plan and the corresponding GT point, and then averages these distances. The detailed formula for the ADE metric used is as follows; 
\begin{equation}
    \text{ADE} = \frac{1}{NF} \sum_{i=1}^N \sum_{j=1}^F \|\textbf{x}_i^{j} - \textbf{g}_i^{j}\|       
\end{equation}
\noindent where $N$ is the number of samples, $F$ is the number of future timesteps, $\mathbf{x}$ is the generated plan, and $\mathbf{g}$ is the ground truth plan.

\item \textbf{Collision Rate (CR)}

Collision Rate is an important metric for evaluating safety in Motion Planning. It considers a collision to occur when the distance between certain waypoints in the generated plan and the ground truth plan is below a specified threshold. Following the approach in GameTheoretic~\cite{kedia2023GameTheoretic}, we use a threshold of 0.6. The formula for Collision Rate is as follows: 
\begin{equation}
     \text{CR} = \frac{1}{N} \sum_{i=1}^{N} \mathbf{1}\left( \min_{j=1,\dots,F} \|\mathbf{x}_i^{j} - \mathbf{g}_i^{j}\| < {\epsilon}_{c} \right)
\end{equation}
\noindent where ${\epsilon}_{c}$ is the collision threshold.
    
    \item \textbf{Final Displacement Error (FDE)}

FDE is a metric for evaluating goal success. It calculates the L2 distance between the position at the final time step of the generated ego agent's plan and the destination. The formula for the FDE metric is as follows:
\begin{equation}
    \text{FDE} = \frac{1}{N} \sum_{i=1}^N \|\textbf{x}_i^{T_f} - \textbf{g}_i^{T_f}\| 
\end{equation}
\noindent where $T_f$ is the final timestep.
    
\item \textbf{Miss Rate (MR)}

Miss rate is also a metric for evaluating goal success, assessing whether the position at the final time step of the generated plan deviates from the destination by more than a specified threshold. We compute the L2 distance between the endpoint of the plan and the destination, and here we use a threshold of 0.5 to determine a miss. The detailed formula for Miss Rate is as follows: 
\begin{equation}
    \text{MR} = \frac{1}{N} \sum_{i=1}^N \mathbf{1}(\|\textbf{x}_i^{T_f} - \textbf{g}_i^{T_f}\| > {\epsilon}_{m})
\end{equation}
\noindent where ${\epsilon}_{m}$ is the miss threshold.

\end{enumerate}

\begin{table}[t]
\caption{Checkpoint pool $\mathcal{P}$ constructed for each planning model. The checkpoint interval $C$ refers to the extraction interval of intermediate epoch points.}
\vspace{-5pt}
\label{tab:appendix_checkpointpool}
\resizebox{\columnwidth}{!}{%
\begin{tabular}{ll|l}
\hline
\multicolumn{2}{l|}{Model}                      & Checkpoint Pool $\mathcal{P}$                                             \\ \hline
\multicolumn{2}{l|}{GameTheoretic (forecaster)} & ADE, $C$ = 30                                             \\
\multicolumn{2}{l|}{GameTheoretic (planner)}    & ADE, CR, FDE, MR, $C$ = 5                              \\
\multicolumn{2}{l|}{DTPP}                       & ADE, CR, FDE, MR, $C$ = 1          \\
\multicolumn{2}{l|}{DIPP}                       & ADE, CR, FDE, MR, $C$ = 5  \\ \hline
\end{tabular}%
}
\end{table}

\section{Implementation Details}
\label{sup:imple}
\begin{enumerate}
    \item \textbf{Interaction-Conserving Pre-Merging}

We extract key checkpoints from each of the three trained planning baseline models. Specifically, we select the checkpoints where ADE, CR, FDE, and MR achieve their best values and store them in the checkpoint pool $\mathcal{P}$. The selection of intermediate checkpoints during training is determined through multiple experiments.
For GameTheoretic~\cite{kedia2023GameTheoretic}, unlike DTPP~\cite{huang2024DTPP} and DIPP~\cite{huang2023DIPP}, a separate forecaster is used. Therefore, we also extract forecaster checkpoints following a similar methodology as the planner. The checkpoints used for each planning model are in Tab.~\ref{tab:appendix_checkpointpool}

    \item \textbf{Interaction Transfer with Merging}

Checkpoints in the checkpoint pool $\mathcal{P}$ are separated by module within the planning model. Specifically, we extract weights from the LSTM layers responsible for embedding the trajectories of the ego agent and surrounding agents, denoted as $\theta_{ego}$ and $\theta_{surr}$, respectively. Additionally, we obtain $\theta_{inter}$ from the transformer model, which embeds interactions between the ego agent and other agents. Other parameters, excluding those related to embedding, are consolidated and stored together.
When merging trainable parameters, we update them based on the loss of the target domain in the original planning model. We use Adam as the optimizer and set the learning rates to 1e-3, 1e-2, and 1e-3 for GameTheoretic, DTPP, and DIPP, respectively. For the scheduler, we use ReduceLROnPlateau across all three models.

\end{enumerate}

\section{Qualitative results}

In Fig.3 of the main paper, the correlation between the merging weights of IMMP and the inference results of models trained on each source domain is shown when the SIT dataset is used as the target domain in the GameTheoretic model. Figure~\ref{fig:viz_qual_THOR_supple} presents the result when the THOR dataset is used as the target domain. In cases~\cref{fig:viz_qual_THOR_supple} (d) and (g), which show poor inference performance, low weights are assigned, while in cases~\cref{fig:viz_qual_THOR_supple} (e) and (f), which show good inference performance, higher weights are assigned. This demonstrates that even when the composition of the target dataset changes, IMMP effectively identifies the important source domains.

\section{Further Analysis}
\label{sup:further_analysis}
\begin{enumerate}
\item \textbf{Need for Metric-wise Checkpoint Collection}
\vspace{1pt}

To demonstrate that metric-wise checkpoints capture distinct characteristics, we evaluated metric-specific checkpoints, selected on the Univ dataset, across two target domains.
As shown in Tab.~\ref{tab:Target_evaluation_of_metric_based_checkpoints_from_Univ}, checkpoints from the same source (Univ) but selected by different metrics (e.g., Collision vs. Miss Rate) behave differently across target domains. Notably, the collision-optimized checkpoint performs best on THOR but worst on SIT, suggesting that each metric captures distinct aspects of the source domain.

\begin{table}[htbp]
\centering
\caption{Target evaluation of metric-wise checkpoints from Univ.}
\vspace{-5pt}
\label{tab:Target_evaluation_of_metric_based_checkpoints_from_Univ}
\resizebox{0.99\columnwidth}{!}{%
\begin{tabular}{c|c|cccc}
\hline
Target                 & Chckpoint & ADE $\downarrow$                            & Col. Rate $\downarrow$                        & FDE $\downarrow$                            & Miss Rate $\downarrow$                      \\ \hline
                       & ADE / FDE & 0.4671                         & 2.54E-04                         & 1.0002                         & 0.8043                         \\
                       & Col. Rate & \cellcolor[HTML]{A4C2F4}0.8023 & \cellcolor[HTML]{A4C2F4}6.08E-04 & \cellcolor[HTML]{A4C2F4}1.6444 & \cellcolor[HTML]{A4C2F4}0.9489 \\
\multirow{-3}{*}{SIT}  & Miss Rate & 0.3956                         & 4.29E-04                         & 0.8730                         & 0.7294                         \\ \hline
                       & ADE / FDE & 0.3692                         & 4.98E-04                         & 0.6602                         & 0.7470                         \\
                       & Col. Rate & \cellcolor[HTML]{EA9999}0.2723 & \cellcolor[HTML]{EA9999}4.91E-04 & \cellcolor[HTML]{EA9999}0.5543 & \cellcolor[HTML]{EA9999}0.4481 \\
\multirow{-3}{*}{THOR} & Miss Rate & 0.3674                         & 5.19E-04                         & 0.6541                         & 0.7488                         \\ \hline
\end{tabular}%
}
\end{table}

\begin{table}[htbp]
\centering
\caption{Effect of checkpoint interval $C$ on Gametheoretic model performance in the SIT target domain (without finetuning).}
\vspace{-5pt}
\label{tab:effect_of_checkpoint_interval_C}
\resizebox{0.99\columnwidth}{!}{%
\begin{tabular}{c|cccc}
\hline
C value  & ADE  $\downarrow$  & Col. Rate $\downarrow$ & FDE $\downarrow$    & Miss Rate $\downarrow$ \\ \hline
1        & 0.3234 & 4.69E-05  & 0.7471 & 0.6214    \\
2        & 0.3219 & 4.28E-05  & 0.7424 & 0.6156    \\
3        & 0.3220 & 4.28E-05  & 0.7433 & 0.6144    \\
10       & 0.3243 & 3.62E-05  & 0.7548 & 0.6446    \\ \hline
Ours (5) & 0.3157 & 4.28E-05  & 0.7300 & 0.5934    \\ \hline
\end{tabular}%
}
\end{table}

\vspace{2pt}
\item \textbf{Hyperparameter Sensitivity: Checkpoint Interval}
\vspace{1pt}

As shown in Tab.~\ref{tab:effect_of_checkpoint_interval_C}, when the GameTheoretic model targets the SIT domain, we evaluated performance across different checkpoint intervals $C$. Our method consistently outperforms the best-performing baseline in Tab.~\textcolor{red}{1} (Ensemble-WTA, ADE: 0.3695), even under various settings of $C$. This demonstrates that our model requires minimal manual tuning and offers high practical utility.

\vspace{2pt}
\item \textbf{Ablation Study on Module Separation}
\vspace{1pt}

As shown in Sec.~\ref{sup:Datasets}, the robot motion datasets differ in ego agent type and interaction mechanisms. Our module separation is designed to reflect these characteristics. To validate its effectiveness, we conduct an ablation comparing the original grouping with variants where two of the three modules are merged. As shown in Tab.~\ref{tab:Module_Separation_ablation}, the results confirm that our original grouping is more effective.

\begin{table}[h]
\centering
\caption{Ablation of Module Separation in DIPP on the SIT target domain (A: Robot, B: Human, C: Interaction)}
\vspace{-5pt}
\label{tab:Module_Separation_ablation}
\resizebox{0.99\columnwidth}{!}{%
\begin{tabular}{c|cccc}
\hline
Grouping          & ADE $\downarrow$            & Col. Rate   $\downarrow$      & FDE      $\downarrow$       & Miss Rate   $\downarrow$    \\ \hline
A + B             & 0.5666          & \textbf{1.36E-04} & 1.0744          & 0.8351          \\
B + C             & 0.5608          & 2.18E-04          & 1.0631          & 0.8217          \\
C + A             & 0.6108          & 1.83E-04          & 1.1462          & 0.8751          \\ \hline
Seperate (origin) & \textbf{0.5112} & 1.60E-04          & \textbf{0.9358} & \textbf{0.7944} \\ \hline
\end{tabular}%
}
\end{table}

\vspace{2pt}
\item \textbf{Correlation Between Domain Similarity and Merging Weights}
\vspace{1pt}

We estimate the similarity between the source and target domains by measuring zero-shot performance, and visualize the correlation between dataset similarity and the merging weights. As shown in Fig.~\ref{fig:viz_merging_weights_with_respect_to_similarity}, there exists a proportional relationship between the merging weights and the domain similarity, indicating that more similar source domains are assigned higher weights.

\begin{figure*}[h]
\vspace{-5pt}
    \centering
    \includegraphics[width=0.80\linewidth]{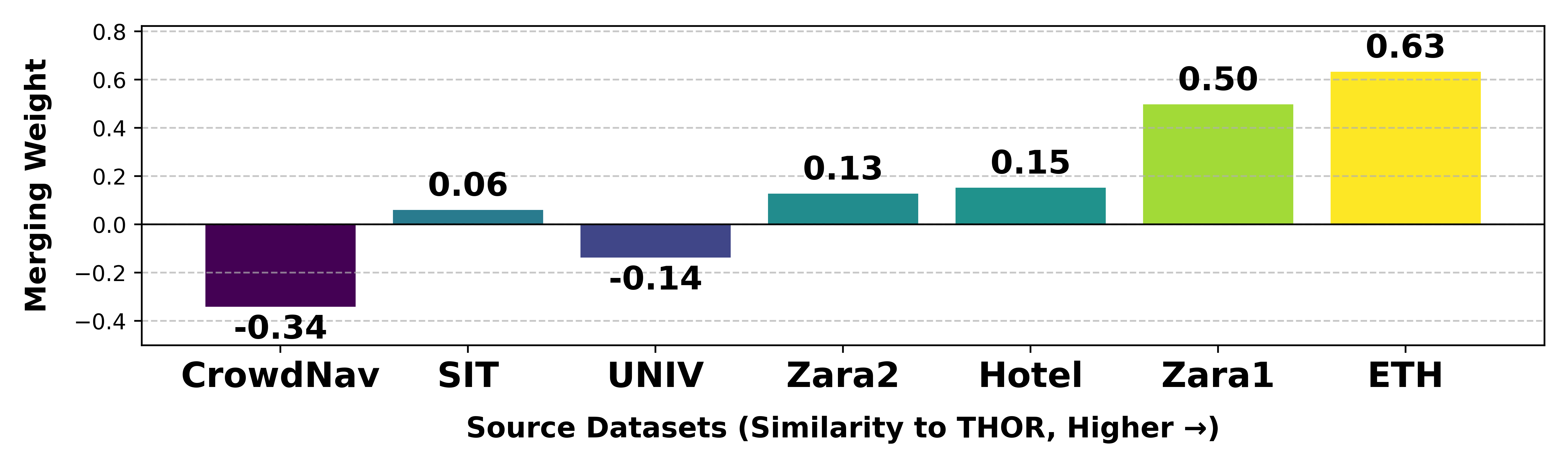}
    \caption{Merging weights with respect to similarity to THOR.}
    \label{fig:viz_merging_weights_with_respect_to_similarity}
\end{figure*}

\vspace{2pt}
\item \textbf{Comparison with Other Merging Techniques}
\vspace{1pt}

We compare with recent merging baselines listed in \cref{tab:merge}, evaluated on the SIT dataset using the GameTheoretic planning model. Existing methods~\cite{Rebuttal_merging1_training_free, Rebuttal_merging2_emr} fail to preserve the hierarchical structure inherent in motion planning, resulting in notable performance degradation. In contrast, IMMP effectively transfers knowledge from source datasets to the target domain, achieving superior performance.

\begin{table}[htbp]
\centering
\vspace{-5pt}
\caption{Comparison with recent merging methods on the SIT target domain.}
\scriptsize
\begin{tabular}{cl|cccc}
\hline
\multicolumn{2}{c|}{Method}   & ADE   $\downarrow$          & Col. Rate    $\downarrow$    & FDE      $\downarrow$       & Miss Rate  $\downarrow$  \\ \hline
\multicolumn{2}{c|}{MuDSC \cite{Rebuttal_merging1_training_free}}          & 1.6729          & 4.21E-03         & 2.9329          & 1.0                \\
\multicolumn{2}{c|}{EMR-Merging \cite{Rebuttal_merging2_emr}}    & 1.5367          & 2.06E-03         & 2.8272          & 0.9941             \\ \hline
\multicolumn{2}{c|}{IMMP}           & \textbf{0.3380} & \textbf{5.12E-05}         & \textbf{0.9580} & \textbf{0.6976}    \\ \hline
\end{tabular}
\label{tab:merge}
\end{table}

\vspace{2pt}
\item \textbf{Experiments on a Larger Target Domain}
\vspace{1pt}

We selected SIT and THOR as target domains because they are real-world datasets collected in actual robotic navigation environments, where data collection is relatively challenging. Such environments are likely to serve as realistic target domains in practical applications. We further evaluate IMMP on Zara2 (3× larger than THOR) in Tab.~\ref{tab:R1_Qminer3_large_scale_targetdomain}, proving robust on large-scale domains.

\begin{table}[htbp]

\centering
\caption{IMMP performance on the Zara2 target domain}
\vspace{-5pt}
\label{tab:R1_Qminer3_large_scale_targetdomain}
\resizebox{0.99\columnwidth}{!}{%
\begin{tabular}{cl|ccccc}
\hline
\multicolumn{2}{c|}{Method}                         & ADE   $\downarrow$          & Col. Rate    $\downarrow$    & FDE      $\downarrow$       & Miss Rate  $\downarrow$     & Cost $\downarrow$ \\ \hline
\multicolumn{2}{c|}{Domain Generalization \cite{gilles2022uncertaintyDomainGeneralization}} & 0.5350          & 0.01802          & 1.1291          & 0.7641          & X1   \\
\multicolumn{2}{c|}{Domain Adaptation \cite{feng2024unitrajDomainGeneralization}}     & 0.5378          & 0.01657          & 1.1441          & 0.7561          & X1   \\
\multicolumn{2}{c|}{Target Only}                    & 0.4850          & 0.01663          & 1.0047          & 0.7224          & X1   \\
\multicolumn{2}{c|}{Ensemble-WTA \cite{arpit2022ensemble}}           & 0.4781          & \textbf{0.01614} & 0.9917          & 0.7078          & X7   \\ \hline
\multicolumn{2}{c|}{IMMP + Finetune}                & \textbf{0.4687} & 0.01688          & \textbf{0.9580} & \textbf{0.6976} & X1   \\ \hline
\end{tabular}%
}
\end{table}

\end{enumerate}

\end{document}